\documentclass[journal]{IEEEtran}

\usepackage{bm}
\usepackage{amsmath}
\usepackage{mathtools}
\usepackage{cuted}
\usepackage{algorithmic}
\usepackage{algorithm}
\usepackage{cite}
\usepackage{color}
\usepackage{mathrsfs}
\usepackage{amsfonts}
\usepackage{amssymb}
\usepackage{makecell}
\usepackage{multirow}
\usepackage{times}
\usepackage{makecell}
\usepackage{tabulary}
\usepackage{array}
\usepackage{url}
\usepackage[table]{xcolor}
\usepackage{colortbl}  
\usepackage{float}
\usepackage{balance}
\usepackage{MnSymbol}
\usepackage{setspace}
\usepackage{booktabs}
\usepackage[normalem]{ulem}
\usepackage{tikz}
\usepackage{xcolor}
\usepackage{graphicx}
\usepackage{booktabs}
\usepackage{diagbox}
\usepackage[flushleft]{threeparttable}
\definecolor{new-red}{rgb}{1.0, 0.5804, 0.5882} 
\definecolor{new-orange}{rgb}{1.0, 0.7843, 0.5725} 
\definecolor{new-yellow}{rgb}{1.0, 0.9647, 0.6627} 

\ifCLASSINFOpdf
\else
\fi

\begin{document}
%
\title{\huge
LOGOS: LiDAR-Only Gaussian Elevation Splatting \\
for Unified Tiny Obstacle Segmentation
}
%
%
%

\author{Nan Ming, Yeqiang Qian*, \textit{Member, IEEE}, Chunxiang Wang, \textit{Member, IEEE}, \\and Ming Yang*, \textit{Member, IEEE}
\thanks{Nan Ming, Yeqiang Qian, Chunxiang Wang, and Ming Yang are with the School of Automation and Intelligent Sensing, Shanghai Jiao Tong University, Shanghai, 200240, China; Key Laboratory of System Control and Information Processing, Ministry of Education, Shanghai, 200240, China.}
\thanks{Corresponding author: Yeqiang Qian (qianyeqiang@sjtu.edu.cn) and Ming
Yang (mingyang@sjtu.edu.cn).}
}

%



\maketitle

\begin{abstract}
Robust obstacle segmentation is essential for the safety of intelligent robots, where LiDAR-based perception systems play a fundamental role in the robot-environment interaction. While extensive LiDAR-based approaches have demonstrated high performance on common obstacles in urban scenarios, their results on tiny obstacles such as curbs, gravel, and potholes remain unsatisfactory due to the significant similarity between tiny obstacles and inherent road undulations. Moreover, their segmentation accuracy even deteriorates sharply when the LiDAR scans suffer from degradation in challenging off-road scenes. To overcome these bottlenecks, we propose LOGOS, a LiDAR-only unified tiny obstacle segmentation system, which models the road surface as a continuous mixture of 2D Gaussian primitives and distinguishes tiny obstacles via high-presicion elevation estimation. Unlike existing Gaussian splatting methods that rely on iterative RGB training, LOGOS is a backpropagation-free LiDAR-only approach. It directly estimates Gaussian parameters via a freespace-aware initialization by incrementally pruning non-road primitives using smoothness constraints. Subsequently, pointwise signed distances are computed via a novel normal-aware elevation splatting function, ensuring robustness to both flat and sloped terrains. We evaluate LOGOS on a highly heterogeneous benchmark of point cloud frames collected from urban mobility scenarios and mining haulage off-road environments. These data are practically acquired using different LiDAR sensors and exhibit large variations in point density, terrain roughness, and obstacle types. Experiments on the road and off-road scenes demonstrate that LOGOS significantly outperforms other state-of-the-art methods, particularly in degraded point cloud regions and challenging off-road scenarios, while maintaining real-time efficiency. 

\end{abstract}

%
\IEEEpeerreviewmaketitle


\section{Introduction}
\label{sec:intro}

\IEEEPARstart{O}{bstacle} perception is crucial to the safe operation of various autonomous robots operating in both structured and unstructured environments, including ground vehicles, legged robots, or industrial platforms \cite{hls}. LiDAR sensors, capable of capturing accurate 3D geometry, have been widely adopted for obstacle segmentation in robotic systems. Numerous LiDAR-based methods \cite{lidarreview} have achieved excellent performance in recognizing common obstacles such as cars, pedestrians, vegetation, and buildings. However, their accuracy degrades sharply when faced with tiny obstacles of small size or low height, including gravel, potholes, curbs, and ruts \cite{small}. This difficulty arises from their tiny size and high geometric similarity to the surrounding terrain, which makes them easily confounded with the intrinsic undulations of the road surface. However, these tiny obstacles pose significant risks to robotic platforms. Positive obstacles (protruding upward) may damage mechanical components or impede locomotion, while negative obstacles (depressed downward) can cause rollover incidents or trap robots in cavities \cite{undercarriage, inconseg}. Therefore, accurate and robust segmentation of tiny obstacles is essential to reduce the risk of mission failure and to enhance the autonomy of robotic systems.

\begin{figure}[!t]  
		\centering
		\includegraphics[width=0.49\textwidth]{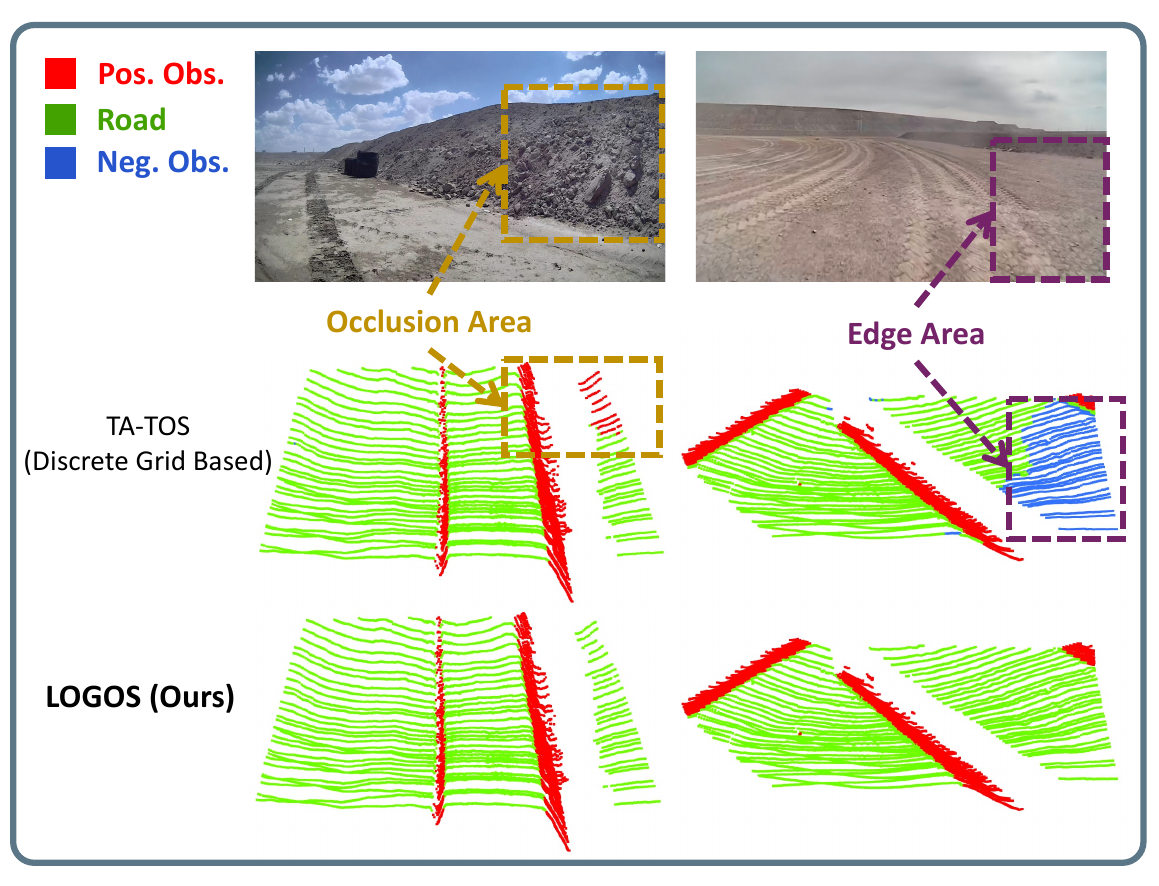}
		\caption{An example of typical degraded LiDAR scans arised from occlusions, or edge degradations. The proposed LOGOS method significantly improves the segmentation accuracy near degraded areas. \protect\textcolor[RGB]{191, 144, 0}{\rule[0ex]{0.8em}{0.8em}} and \protect\textcolor[RGB]{116, 35, 104}{\rule[0ex]{0.8em}{0.8em}} boxes refer to the occlusion and edge areas, respectively.
}
		\label{fig:front}
\end{figure}

For common obstacle segmentation, numerous state-of-the-art (SOTA) classification-based methods attempt to directly classify points into predefined obstacle categories using preset or pretrained features. Although they perform well in detecting common large obstacles, they become struggling when encountering tiny obstacles that are easily blended into road undulations, especially in complex off-road scenes. This difficulty is further compounded by the long-tail distribution of tiny obstacles, where many rare types appear infrequently, making it impractical to collect sufficient training samples for all categories in various operating scenarios. 

Consequently, a series of road-modeling-based approaches have been proposed specifically for tiny obstacles. They first reconstruct a smooth road surface and then segment obstacles via signed distance. For efficiency, most SOTA methods assume either a fixed parametric model (e.g., plane or quadric) or a discrete grid representation. Fixed parametric models inherently lack the flexibility to fit complex undulating terrains, leading to large errors on bumpy roads. Discrete grid representations, though more flexible, impose a piecewise-constant approximation of the continuous surface critically depending on sufficient point density within each grid cell. In degraded LiDAR scans caused by long sensing distances, occlusions, or irregular point distributions at scene edges (as illustrated in Fig.~\ref{fig:front}), grid-based methods suffer from unreliable local estimation due to insufficient inlier points. Therefore, both types exhibit significant performance degradation when handling diverse real-world scenarios with varying terrain undulations and point cloud densities.

To overcome the accuracy limitations of fixed parametric models and discrete grid representations in degraded LiDAR scans, we propose LOGOS, a LiDAR-only Gaussian-splatting-based tiny obstacle segmentation system. Inspired by recent advances in 3D Gaussian Splatting \cite{3dgs} and 2D Gaussian Splatting \cite{2dgs}, LOGOS treats the road surface as a mixture of 2D Gaussian primitives. However, unlike existing Gaussian splatting methods that rely on iterative gradient-based training on RGB images, LOGOS is specifically designed for LiDAR point clouds with a backpropagation-free pipeline. Our method directly estimates Gaussian parameters (mean, covariance, and normal) from local point cloud statistics using a freespace-aware pruning strategy. During initialization, it incrementally rejects Gaussians that violate the smoothness constraint, ensuring that only road-surface Gaussians are retained. In the elevation splatting stage, a novel normal-aware splatting function is then proposed to blend contributions from nearby Gaussians weighted by horizontal Mahalanobis distance. The proposed method is evaluated on highly heterogeneous point clouds collected from two distinct transportation systems, including urban mobility and mining haulage scenarios. The main contributions of this paper are summarized as follows:

\begin{itemize}
    \item [1)] We propose \textbf{LOGOS}, a \textbf{L}iDAR-\textbf{O}nly \textbf{G}aussian-splatting-based tiny \textbf{O}bstacle \textbf{S}egmentation system, which introduces a Gaussian road surface representation to overcome the accuracy limitations of fixed parametric or discrete grid models in degraded LiDAR scans in occlusion or edge areas, enabling real-time performance at the same time.
    \item [2)] We propose the first LiDAR-only \textbf{Gaussian Elevation Splatting} method specifically for point cloud input, which computes pointwise signed distances using a novel \textbf{Normal-Aware Elevation Splatting Function}. It achieves high accuracy on both flat and sloped terrains by leveraging the continuous representation and adaptive density handling of Gaussian splatting.
    \item [3)] We develop a \textbf{Freespace-Aware Initialization} scheme, which jointly estimate and prune road-surface Gaussians without backpropagation training, ensuring both road modeling accuracy and computational efficiency.
    \item [4)] We conduct comprehensive experiments both in urban mobility systems and mining haulage systems, where highly heterogeneous point clouds of varying density from road and off-road scenes are collected using different types of LiDAR.
\end{itemize}

The remainder of this paper is organized as follows. Section~\ref{sec:related-work} reviews related work on tiny obstacle segmentation. Section~\ref{sec:meth} details the proposed LOGOS methodology. Section~\ref{sec:exp} presents experimental results and analysis. Section~\ref{sec:con} concludes the paper.

\section{Related Work}
\label{sec:related-work}

In this section, we first review state-of-the-art (SOTA) tiny obstacle segmentation methods, which can be broadly divided into two categories: classification-based approaches and road-modeling-based approaches. To further improve their accuracy near degraded areas, we then review advanced continuous surface representations, which have the potential to be referential.

\subsection{Classification-Based Methods}
\label{sec:classification}

Early classification-based methods rely on geometric priors and hand-crafted features without requiring training data. For instance, Rahmani \textit{et al.} \cite{rahmani2019gravel} estimated gravel particle size and weight for segmentation, while Du \textit{et al.} \cite{review} applied K-means clustering and region growing for pothole extraction. However, these methods are often iterative and inefficient for scenes with numerous tiny obstacles. More general ground segmentation approaches, such as PMF-GroundSeg \cite{pmf}, employ progressive morphological filters to separate ground from non-ground points, but they still lack fine-grained distinction of small objects.

With the advent of deep learning, a series of Lidar-based semantic segmentation methods (e.g., RangeNet++ \cite{rangenet++}, 2DPASS \cite{2dpass}, PVKD \cite{pvkd}, SphereFormer \cite{sphere}, TransRVNet \cite{transrvnet}, TFNet \cite{tfnet}, TASeg \cite{taseg}) treat each point or voxel as a class (e.g., road, car, pedestrian). Although they achieve high accuracy on common large obstacles, they face two major challenges for tiny obstacle segmentation. First, tiny obstacles occupy very few points and their geometric or semantic features are easily confounded with natural terrain undulations. As a result, these methods tend to classify low-height objects such as curbs, small rocks, or shallow potholes as part of the ground, failing to provide the required fine-grained segmentation. Second, the long-tail distribution of tiny obstacles means that many rare categories lack sufficient training samples, leading to significant accuracy degradation. Moreover, most of these deep learning methods are computationally intensive and struggle to meet the real-time requirements of resource-constrained autonomous robots.

\subsection{Road-Modeling-Based Methods}
\label{sec:road-modeling}

Road-modeling-based approaches focus on reconstructing a smooth road surface and then segmenting tiny obstacles via signed distance. According to the underlying surface representation, they can be further divided into fixed parametric models, discrete grid models, and learned regression models.

\subsubsection{Fixed Parametric Models}

Dhiman \textit{et al.} \cite{pm} modeled the road as a planar manifold using RANSAC \cite{ransac_algo}. Wu \textit{et al.} \cite{qm} extended this to a quadric surface for better representation of curved roads. While these parametric models are computationally efficient, they inherently lack the flexibility to fit complex undulating terrains. In off-road scenes such as mining areas, they often mistake natural terrain variations for obstacles, causing numerous false positives and making them impractical for real-world deployment.

\subsubsection{Discrete Grid Models}

To handle uneven terrains, Patchwork \cite{patchwork} and its extension Patchwork++ \cite{patchwork++} adopt a multi-region ground plane fitting strategy based on principal component analysis (PCA). DipG-Seg \cite{dipg} integrates features from multiple receptive fields to enhance ground recognition. More recently, TA-TOS \cite{tatos} introduced a non-parametric Markov random field (MRF) based road modeling approach, which constructs a grid-based height map and refines it by minimizing a novel negative exponential energy function. However, discrete grid models suffer from a fundamental limitation: their accuracy heavily depends on sufficient point density within each grid cell. In degraded LiDAR scans caused by long sensing distances, occlusions, or irregular point distributions at scene edges, point density can become extremely sparse. Under such conditions, the RANSAC-based initial height estimation within each grid becomes unreliable, and subsequent refinements cannot fully recover the errors. Consequently, the performance of grid-based methods collapses in low-density regions.

\subsubsection{Learned Regression Models}

Direct regression methods learn to predict a continuous height map of the road surface. For example, GndNet \cite{gndnet} uses a neural network to regress a grid-based elevation map directly from LiDAR point clouds, and then classifies obstacles by thresholding the height difference. GndNet achieves real-time inference by avoiding per-point classification. However, its accuracy is limited by the fixed grid resolution and the quality of training data. Moreover, it does not explicitly model the smoothness prior of the road surface, leading to noisy elevation estimates in sparse regions. The long-tail issue also affects regression models, as sparse or rare terrain patterns are underrepresented in training sets.

Therefore, existing road-modeling-based methods struggle to maintain accuracy and robustness under varying point cloud densities, sensor characteristics, and complex off-road terrains. The root cause lies in the limited representational capacity of fixed parametric models and discrete grid representations, which fail to capture the subtle differences between tiny obstacles and natural terrain variations, especially in degraded LiDAR scans or bumpy off-road scenarios. This calls for a more accurate and computationally efficient surface representation that can adapt to diverse real-world conditions.

\subsection{Advanced Continuous Surface Representations}

To achieve high fidelity modeling of complex surfaces, the fields of computer graphics and 3D vision have developed three main families of methods: mesh based, signed distance function (SDF) based, and Gaussian splatting based approaches. We review each of them in the context of tiny obstacle segmentation.

Mesh based methods explicitly represent surface geometry through vertices and facets. This family of methods has a long history and mature techniques, and can accurately describe complex topologies. For example, \cite{shape} proposed a multi-resolution mesh representation that efficiently models 3D shapes at different levels of detail, while \cite{grid} studied deformable mesh models for non-rigid object reconstruction. In autonomous driving, \cite{squeezeseg} applied mesh models to road surface reconstruction, but their high computational complexity limits real-time performance. \cite{kongsemantic} further proposed a LiDAR based road mesh generation algorithm, which also faces real-time challenges. Although mesh models offer high accuracy, their discrete facet structure requires substantial memory and computational resources for optimization and rendering. Moreover, they are sensitive to point cloud noise and missing data.

Signed distance function based methods represent surfaces through continuous implicit functions whose zero level sets correspond to the surface. These methods naturally capture complex topologies and fine geometric details. A seminal work is DeepSDF \cite{sdf}, which uses a deep network to learn a mapping from 3D coordinates to signed distances, achieving high quality shape representation. Neural radiance fields (NeRF) \cite{nerf} are primarily designed for view synthesis with limited geometric modeling capability, yet their implicit representation philosophy has profoundly influenced subsequent SDF research. NeuS style methods improve volume rendering to enhance surface modeling, but they rely on neural networks for distance field estimation and suffer from low efficiency \cite{neus2}. \cite{instant} introduced efficient hash encoding to greatly accelerate the training of NeRF like implicit models, enabling outdoor scene reconstruction, but still falling short of real time inference. Overall, SDF methods can achieve high accuracy surface reconstruction, but their dependence on large neural networks for inference or optimization incurs huge computational costs and lacks real time potential.

Gaussian splatting based methods have recently emerged as a promising paradigm. They represent a scene using a set of Gaussian primitives, each with attributes such as position, covariance, and opacity. Compared to mesh and SDF approaches, Gaussian splatting achieves a better balance between rendering speed and reconstruction quality. For instance, 3D Gaussian Splatting (3DGS) \cite{3dgs} enables high quality real time scene reconstruction and rendering. Event 3DGS \cite{event} further improves training and rendering speeds, demonstrating real time potential. In particular, for scenes rich in planar features with small curvature variations, such as road surfaces, 2D Gaussian Splatting (2DGS) \cite{2dgs} defines Gaussians on 2D planes, significantly enhancing modeling efficiency and accuracy. The differentiable rendering pipeline, flexible scene representation, and relatively low computational overhead of Gaussian splatting make it a promising foundation for tiny obstacle segmentation from LiDAR point clouds.

However, directly applying these Gaussian splatting frameworks to the tiny obstacle segmentation task poses three fundamental challenges. First, traditional Gaussian Splatting takes multi-view RGB images as input, while our system relies solely on 3D LiDAR point clouds, which provide accurate geometric information but lack color or texture cues. Furthermore, conventional Gaussian Splatting methods require iterative gradient-based optimization to refine Gaussian parameters, which is computationally expensive and unsuitable for real-time perception in autonomous transportation systems. Finally, the goal of traditional methods is to generate photorealistic renderings, whereas our objective is to obtain accurate elevation estimates for each 3D point and subsequently perform precise obstacle segmentation.

\begin{figure*}[!t]  
		\centering
		\includegraphics[width=1\textwidth]{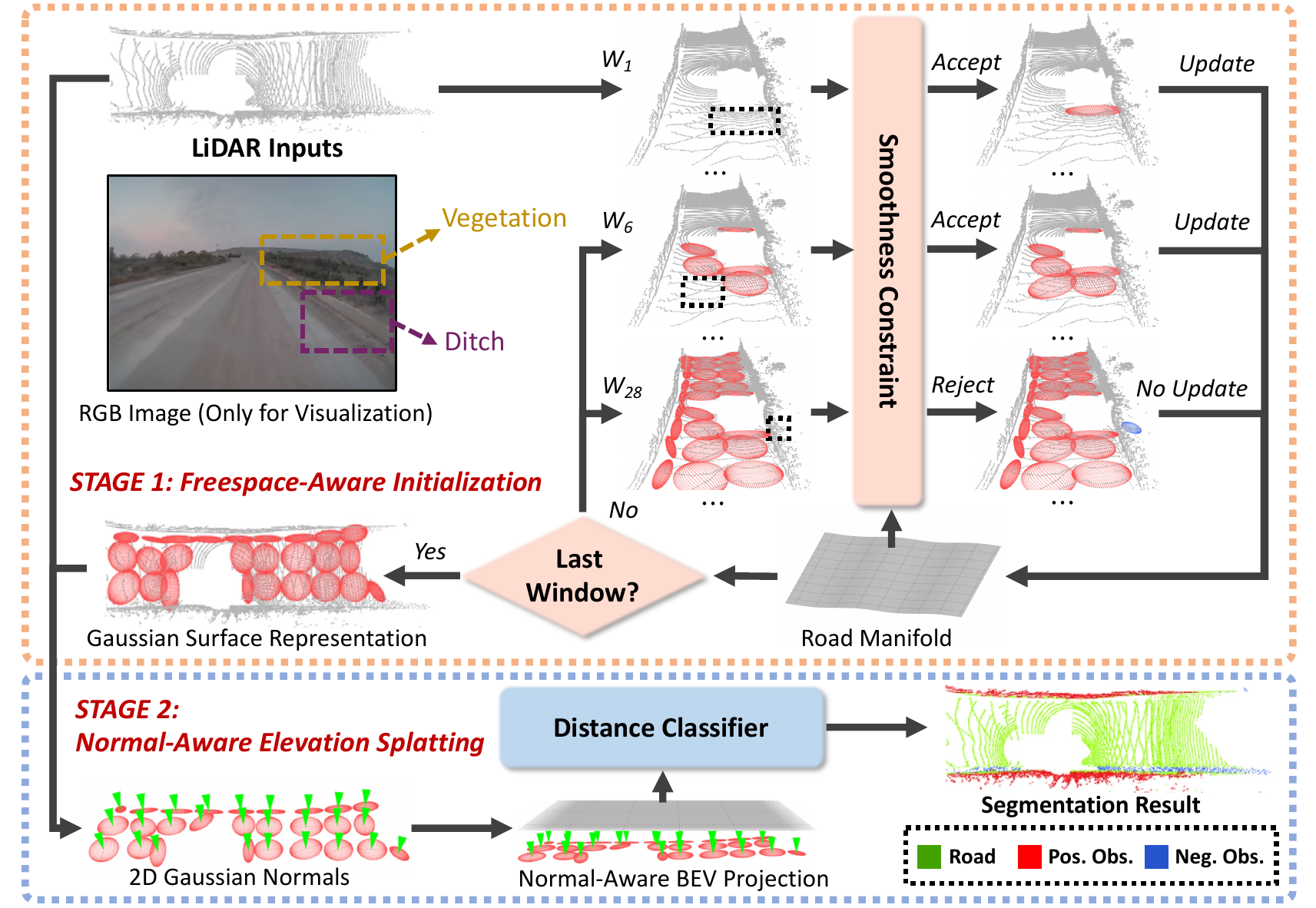}
		\caption{An illustration of our proposed LOGOS method, including freespace-aware initialization stage and normal-aware elevation splatting stage. A sliding-window strategy directly estimates Gaussian primitives from local point cloud statistics. An incremental pruning strategy, guided by terrain smoothness, filters out obstacle or noise Gaussians to retain only road-surface primitives. Subsequently, a normal-aware BEV projection module computes pointwise signed distances using Gaussian normals, and thresholding these distances yields the final segmentation. }
		\label{fig:meth}
\end{figure*}

\section{Methodology}
\label{sec:meth}

\subsection{System Overview}
As discussed in Sec.~\ref{sec:related-work}, although Gaussian splatting methods have emerged as powerful paradigms for 3D scene representation \cite{3dgs, 2dgs, event, instant} and promise to improve terrain reconstruction accuracy in off-road or degraded scenarios, applying these frameworks directly to the tiny obstacle segmentation task presents several challenges. To address these, we propose \textbf{LOGOS}, a novel framework that migrates Gaussian Splatting for terrain-aware tiny obstacle segmentation to avoid the accuracy degradation in off-road scenes or challenging regions. 

As illustrated in Fig.~\ref{fig:meth}, our method introduces three key modules. First, we propose a \textbf{Freespace-Aware Initialization} scheme. This strategy directly estimates Gaussian parameters (mean, covariance, and normal) from local point cloud statistics using a sliding-window scheme, while incrementally pruning Gaussians that violate terrain smoothness constraints. By eliminating backpropagation process in the original Gaussian splatting advances, it ensures both modeling accuracy and computational efficiency. In addition, we develop a \textbf{Normal-Aware Elevation Splatting} module, which computes pointwise signed distances to the Gaussian-represented surface without relying on a camera projection center. This function blends contributions from nearby Gaussians weighted by the horizontal Mahalanobis distance in the bird's-eye-view (BEV) plane, with the local projection orientation determined by each Gaussian's normal. Finally, the segmentation is obtained by thresholding these signed distances to classify points as road, positive obstacles, or negative obstacles. These modules enable LOGOS to handle diverse terrains and challenging point distributions without requiring RGB data or iterative training.

\subsection{Gaussian Primitives Initialization}
\label{sec:meth:init}

The first step of LOGOS is to construct a Gaussian Mixture Model (GMM) representation of the road surface from raw LiDAR point clouds. Unlike conventional GMM estimation that relies on iterative expectation-maximization, we propose a sliding-window-based initialization that directly computes Gaussian parameters from local point cloud statistics, enabling efficient and deterministic initialization.

\subsubsection{Sliding-Window Strategy}

Given an input point cloud $\mathcal{P} = \{\mathbf{p}_i\}_{i=1}^{N} \subset \mathbb{R}^3$, we first define a rectangular region of interest (ROI) in the birds's eye view (BEV) plane, i.e., X-Y plane, covering the area around the ground vehicles. To generate Gaussian primitives within this ROI, we employ a sliding-window approach.

Denote the window size as $W \times W$ (in meters) and the stride as $S$ (in meters). Typically, we set $S < W$ to ensure overlap between adjacent windows, preventing modeling gaps at window boundaries. Starting from a high-density region near the truck, we sequentially generate window centers $\{\mathbf{w}_j\}_{j=1}^{M}$, where $M$ depends on the ROI size and stride.

For the $j$-th window centered at $\mathbf{w}_j$, we extract the point subset $\mathcal{P}_j$ that falls within this window:
\begin{equation}
    \mathcal{P}_j = \left\{ \mathbf{p} \in \mathcal{P} \mid |p_x - w_{j,x}| \le W/2, \; |p_y - w_{j,y}| \le W/2 \right\},
    \label{eq:window_crop}
\end{equation}
where $p_x, p_y$ denote the X and Y coordinates of point $\mathbf{p}$, and $w_{j,x}, w_{j,y}$ are the coordinates of the window center.

We then define a minimum point threshold $N_{\min}$. If $|\mathcal{P}_j| < N_{\min}$, the window is considered too sparse for reliable parameter estimation and is skipped. This strategy allows our method to adaptively focus on regions with sufficient point cloud support, ignoring areas with too sparse LiDAR scans, especially in distant or heavily occluded regions.

\subsubsection{Maximum Likelihood Estimation of Gaussian Parameters}

For windows containing sufficient points ($|\mathcal{P}_j| \ge N_{\min}$), we assume these points are sampled from a 3D Gaussian distribution and estimate its parameters via maximum likelihood estimation (MLE).

The mean vector $\mathbf{g}_j \in \mathbb{R}^3$ is computed as the arithmetic mean of all points in the window:
\begin{equation}
    \mathbf{g}_j = \frac{1}{K} \sum_{k=1}^{K} \mathbf{p}_k,
    \label{eq:mean}
\end{equation}
where $K = |\mathcal{P}_j|$.

The covariance matrix $\mathbf{G}_j \in \mathbb{R}^{3 \times 3}$ is estimated as:
\begin{equation}
    \mathbf{G}_j = \frac{1}{K} \sum_{k=1}^{K} (\mathbf{p}_k - \mathbf{g}_j)(\mathbf{p}_k - \mathbf{g}_j)^\top.
    \label{eq:cov}
\end{equation}

Since the road surface is intrinsically a 2D manifold, the point distribution within a window should be elongated along the tangent plane and highly compact along the normal direction. To better capture this property and improve numerical stability, we enforce the Gaussian to be 2D by setting its smallest eigenvalue to zero. Specifically, we perform singular value decomposition (SVD) on $\mathbf{G}_j$:
\begin{equation}
    \mathbf{G}_j = \mathbf{U} \operatorname{diag}(\sigma_1, \sigma_2, \sigma_3) \mathbf{V}^\top,
    \label{eq:svd}
\end{equation}
with $\sigma_1 \ge \sigma_2 \ge \sigma_3 \ge 0$. We then set $\sigma_3 = 0$ to obtain a rank-2 covariance matrix:
\begin{equation}
    \mathbf{G}_j' = \mathbf{U} \operatorname{diag}(\sigma_1, \sigma_2, 0) \mathbf{V}^\top.
    \label{eq:rank2_cov}
\end{equation}

The third column of the right singular matrix $\mathbf{V}$, corresponding to the smallest singular value, gives the normal vector of the local plane. We extract this as each Gaussian normal $\mathbf{n}_j$, and ensure consistent orientation by flipping it to point upward the positive Z component:
\begin{equation}
    \mathbf{n}_j = \mathrm{sgn}(n_{j,z}) \cdot \frac{\mathbf{v}_3}{\|\mathbf{v}_3\|}.
    \label{eq:normal}
\end{equation}

At this stage, we have obtained a set of initial Gaussian primitives $\{(\mathbf{g}_j, \mathbf{G}_j', \mathbf{n}_j)\}_{j=1}^{\tilde{M}}$, where $\tilde{M}$ is the number of valid windows. However, this initial set includes Gaussians generated from windows that may contain obstacles or noise. These must be filtered to retain only those that truly represent the road surface.

\begin{algorithm}[t!]
    \caption{Freespace-Aware Gaussian Initialization}
    \label{alg:pruning}
    \begin{algorithmic}[1]
        \REQUIRE Point cloud $\mathcal{P}$, ordered windows $\{\mathbf{w}_j\}_{j=1}^{M}$, smoothness threshold $\epsilon_c$, minimum point threshold $N_{\min}$
        \ENSURE Gaussian means $\bm{\mu} = \{\mathbf{g}_j\}$, covariances $\bm{\Sigma} = \{\mathbf{G}_j\}$, normals $\bm{N} = \{\mathbf{n}_j\}$
        
        \STATE $\bm{\mu} \leftarrow \emptyset,\ \bm{\Sigma} \leftarrow \emptyset,\ \bm{N} \leftarrow \emptyset$
        \FOR{each window $\mathbf{w}_j$ in $\mathcal{W}$}
            \STATE $\bm{P} \leftarrow \operatorname{crop}(\mathcal{P}, \mathbf{w}_j)$
            \STATE $\mathbf{g} \leftarrow \operatorname{mean}(\mathbf{p} \in \bm{P})$
            \IF{$\mathcal{L}_{\text{smoothness}}(\mathbf{g}, \bm{\mu}) < \epsilon_c$}
                \STATE \textbf{continue} \hfill \textit{// Reject outlier Gaussian}
            \ENDIF
            \STATE $\mathbf{G} \leftarrow \operatorname{cov}(\mathbf{p} \in \bm{P})$
            \STATE $[\mathbf{U},\ \operatorname{diag}(\bm{\sigma}),\ \mathbf{V}] \leftarrow \operatorname{svd}(\mathbf{G})$
            \STATE $\sigma_3 \leftarrow 0$
            \STATE $\mathbf{G} \leftarrow \mathbf{U} \operatorname{diag}(\bm{\sigma}) \mathbf{V}^\top$
            \STATE $\mathbf{n} \leftarrow \mathbf{V}[:, 3]$
            \STATE $\mathbf{n} \leftarrow \mathrm{sgn}(\mathbf{n}_z) \cdot \mathbf{n} / \|\mathbf{n}\|$
            \STATE $\bm{\mu} \leftarrow \bm{\mu} \cup \{\mathbf{g}\},\ \bm{\Sigma} \leftarrow \bm{\Sigma} \cup \{\mathbf{G}\},\ \bm{N} \leftarrow \bm{N} \cup \{\mathbf{n}\}$
        \ENDFOR
        \RETURN $\bm{\mu},\ \bm{\Sigma},\ \bm{N}$
    \end{algorithmic}
\end{algorithm}

\subsection{Freespace-Aware Pruning for Gaussian Primitives}
\label{sec:meth:pruning}

The initial Gaussian set is over-complete and contains both road and obstacle Gaussians. Directly using all primitives for road modeling would severely distort the reference surface. Therefore, we propose an incremental pruning strategy that leverages the inherent smoothness prior of road surfaces to filter out anomalous Gaussians during initialization.

\subsubsection{Problem Formulation via $\ell_0$ Optimization}

Ideally, we seek an optimal subset of Gaussian means $\bm{\mu}^* = \{\mathbf{g}_j^*\}$ that is as close as possible to the initial estimates $\bm{\mu}_{\text{initial}} = \{\mathbf{g}_j\}$ while satisfying global smoothness constraints. This can be naturally formulated as an $\ell_0$ minimization problem:
\begin{equation}
    \begin{array}{ll}
        \min\limits_{\bm{\mu}} & \|\bm{\mu} - \bm{\mu}_{\text{initial}}\|_0 \\
        \text{s.t.} & \mathcal{L}_{\text{smoothness}}(\bm{\mu}) \le \epsilon_c \cdot \mathbf{1},
    \end{array}
    \label{eq:l0norm}
\end{equation}
where $\|\cdot\|_0$ denotes the $\ell_0$ norm reflecting the number of non-zero elements, $\mathcal{L}_{\text{smoothness}}(\cdot)$ is a function measuring the roughness of the Gaussian set, $\epsilon_c$ is the maximum allowed roughness threshold, and $\mathbf{1}$ is the all-ones vector. This optimization aims to retain as many initial Gaussians as possible while ensuring that the retained set satisfies smoothness constraints. However, solving this $\ell_0$ problem is NP-hard and infeasible for real-time applications.




\begin{figure}[!t]  
		\centering
		\includegraphics[width=0.47\textwidth]{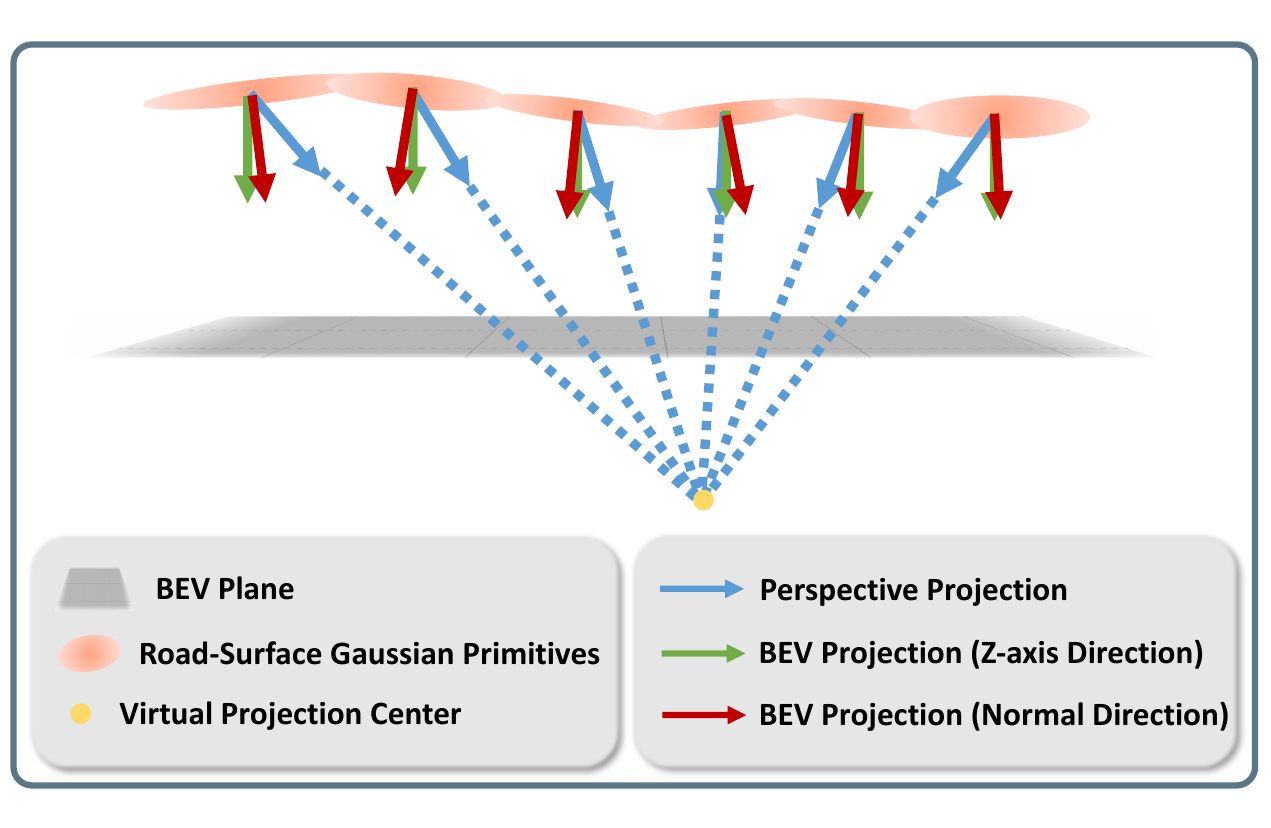}
		\caption{An illustration of three different rendering methods for Gaussian splatting: perspective projection, BEV projection in the Z-axis direction, and BEV projection in the normal direction, respectively.}
		\label{fig:proj}
\end{figure}

\subsubsection{Incremental Gaussian Initialization and Pruning Strategy}

Instead of solving the global $\ell_0$ optimization, we propose an efficient incremental algorithm that processes windows in spatial order and makes pruning decisions immediately. The key insight is that due to spatial continuity, a newly initialized Gaussian should be consistent with already-accepted Gaussians representing the nearby road surface.

Algorithm~\ref{alg:pruning} describes this incremental pruning procedure. We initialize an empty set of accepted Gaussians. For each window processed in spatial order, its Gaussian parameters are computed as described in Section~\ref{sec:meth:init}. We then evaluate the smoothness constraint using the already-accepted Gaussians as reference. If the candidate Gaussian is within the smoothness threshold $\epsilon_c$, the point is sufficiently close to the local surface formed by its neighbors and then accepted; otherwise, it is rejected as an obstacle or noise Gaussian. This local decision-making process has linear complexity and is suitable for real-time deployment.

\subsection{Elevation Splatting Based on BEV Projection}
\label{sec:meth:rendering}

After obtaining a purified set of road-surface Gaussians $\{(\mathbf{g}_j, \mathbf{G}_j, \mathbf{n}_j)\}$, we compute the signed distance of each LiDAR point to the Gaussian-represented surface. This step is analogous to the rendering process in classical Gaussian Splatting but adapted for LiDAR data. We propose a BEV-projection-based rendering that avoids the perspective distortion inherent in traditional projective methods while accounting for local surface orientation via Gaussian normals.

\subsubsection{Limitations of Perspective Projection}

Traditional 3DGS uses perspective projection to render 3D Gaussians onto a 2D image plane, simulating camera imaging. If we attempted to directly apply this concept to elevation rendering, a virtual projection center is necessary. However, LiDAR sensors do not have a single projection center like a pinhole camera; they scan the environment via rotating laser beams. Arbitrarily choosing a projection center leads to the unbalanced spatial resolution, causing near-field regions to dominate the rendering while far-field regions are undersampled. This leads to inconsistent elevation estimates across the ROI, which is unacceptable for autonomous driving applications that require uniform perception accuracy.

\begin{algorithm}[t!]
    \caption{Normal-Aware Elevation Splatting and Segmentation}
    \label{alg:rendering}
    \begin{algorithmic}[1]
        \REQUIRE Point cloud $\mathcal{P}$, Gaussian means $\bm{\mu}=\{\mathbf{g}_j\}$, covariances $\bm{\Sigma}=\{\mathbf{G}_j\}$, normals $\bm{N}=\{\mathbf{n}_j\}$, elevation threshold $\epsilon_d$
        \ENSURE Signed distances $\bm{h} = \{h(\mathbf{p}_i)\}$, labels $\bm{l} = \{l(\mathbf{p}_i)\}$
        
        \STATE $\bm{h} \leftarrow \mathbf{0},\ \bm{l} \leftarrow \bm{\mu}$
        \FOR{$j = 1$ to $M$}
            \STATE $\mathbf{g}_j^{\text{2D}} \leftarrow \mathbf{g}_j[1:2]$
            \STATE $\mathbf{G}_j^{\text{2D}} \leftarrow \mathbf{G}_j[1:2][1:2]$
        \ENDFOR
        \FOR{each point $\mathbf{p}_i \in \mathcal{P}$}
            \STATE $w_i \leftarrow \exp \big\{-(\mathbf{p}_i[1:2]-\mathbf{g}_j^{\text{2D}})^\top\mathbf{G}_j^{\text{2D}}(\mathbf{p}_i[1:2]-\mathbf{g}_j^{\text{2D}})\big\}$
            \STATE $W \leftarrow \mathrm{sum}(w_i)$
            \STATE $h(\mathbf{p}_i) \leftarrow \frac{1}{W}\cdot\mathrm{sum}\big((\mathbf{p}_i-\mathbf{g}_j)^\top \mathbf{n}_j \big) \cdot w_i$
            \IF{$h(\mathbf{p}_i) > \epsilon_d$}
                \STATE $l(\mathbf{p}_i) \leftarrow 1$
            \ELSIF{$h(\mathbf{p}_i) < -\epsilon_d$}
                \STATE $l(\mathbf{p}_i) \leftarrow -1$
            \ELSE
                \STATE $l(\mathbf{p}_i) \leftarrow 0$
            \ENDIF
        \ENDFOR
        \RETURN $\bm{h},\ \bm{l}$
    \end{algorithmic}
    \label{{alg:rendering}}
\end{algorithm}

\subsubsection{BEV Projection with Z-axis Difference}

To overcome these limitations, we adopt a BEV projection scheme. Both 3D points and Gaussians are projected vertically onto the BEV plane, and all computations are performed in the BEV space. The most straightforward approach is to use the Z-coordinate difference for blending:
\begin{equation}
    \begin{split}
    h(\mathbf{p}) & = \frac{1}{W} \sum_{j=1}^{\tilde{M}} \big( z(\mathbf{p}) - z(\mathbf{g}_j) \big) \cdot w_j, \\
    w_j & = \exp\left\{ -(\mathbf{p}^{\text{2D}} - \mathbf{g}_j^{\text{2D}})^\top \mathbf{G}_j^{\text{2D}} (\mathbf{p}^{\text{2D}} - \mathbf{g}_j^{\text{2D}}) \right\},
    \end{split}
    \label{eq:z_rendering}
\end{equation}
where $\mathbf{p}^{\text{2D}} = (p_x, p_y)^\top$, $\mathbf{g}_j^{\text{2D}} = (g_{j,x}, g_{j,y})^\top$ are the BEV projections, and $\mathbf{G}_j^{\text{2D}} \in \mathbb{R}^{2 \times 2}$ is the top-left $2\times 2$ submatrix of $\mathbf{G}_j$, capturing the Gaussian’s spatial extent in the BEV plane. The exponential term is a 2D Gaussian weight based on Mahalanobis distance, and $W = \sum_j w_j$ is the normalization factor.

While this method is simple, it assumes the road is locally horizontal. On sloped terrains, the Z-difference $z(\mathbf{p}) - z(\mathbf{g}_j)$ does not accurately reflect the true distance along the road normal, leading to systematic errors.

\subsubsection{BEV Projection with Normal-Aware Splatting}

To accurately handle sloped surfaces, we leverage the Gaussian normal vectors computed during initialization. The improved elevation rendering function is defined as:
\begin{equation}
    \begin{split}
    h(\mathbf{p}) & = \frac{1}{W} \sum_{j=1}^{\tilde{M}} \big( (\mathbf{p} - \mathbf{g}_j)^\top \mathbf{n}_j \big) \cdot w_j, \\
    w_j & =\exp\left\{ -(\mathbf{p}^{\text{2D}} - \mathbf{g}_j^{\text{2D}})^\top \mathbf{G}_j^{\text{2D}} (\mathbf{p}^{\text{2D}} - \mathbf{g}_j^{\text{2D}}) \right\}.
    \end{split}
    \label{eq:normal_rendering}
\end{equation}

The key improvement is replacing the Z-difference $z(\mathbf{p}) - z(\mathbf{g}_j)$ with the projection of the vector $(\mathbf{p} - \mathbf{g}_j)$ onto the local normal $\mathbf{n}_j$. This quantity $(\mathbf{p} - \mathbf{g}_j)^\top \mathbf{n}_j$ represents the signed distance from point $\mathbf{p}$ to the local tangent plane passing through $\mathbf{g}_j$ with normal $\mathbf{n}_j$. The final signed distance $h(\mathbf{p})$ is a weighted average of these Gaussian signed distances, where weights are determined by the horizontal Mahalanobis distance. This formulation combines the stability of BEV projection with the geometric accuracy of normal-aware distance computation to avoid perspective distortion and to handle terrain slopes, respectively. The illustration of perspective projection, BEV projection in the Z-axis direction, and BEV projection in the Gaussian normal direction is shown in Fig.~\ref{fig:proj}.

\subsubsection{Obstacle Segmentation}

Once the signed distance $h(\mathbf{p})$ is computed for each point, we apply a simple thresholding scheme to obtain the final segmentation labels $l(\mathbf{p}) \in \{0, 1, -1\}$:
\begin{equation}
    l(\mathbf{p}) = 
    \begin{cases}
        1, & \text{if } h(\mathbf{p}) > \epsilon_d \quad \text{(positive obstacle)} \\
        -1, & \text{if } h(\mathbf{p}) < -\epsilon_d \quad \text{(negative obstacle)} \\
        0, & \text{otherwise} \quad \text{(road surface)}
    \end{cases}
    \label{eq:segmentation}
\end{equation}
where $\epsilon_d$ is a pre-set elevation threshold representing the minimum detectable obstacle elevation.

The complete rendering and segmentation procedure is summarized in Algorithm~\ref{alg:rendering}. This algorithm processes each point independently, making it highly parallelizable and suitable for real-time implementation.

\section{Experiment}
\label{sec:exp}

\subsection{Experimental Setup}
\subsubsection{Dataset Description}
\label{sec:dataset}

\begin{figure*}[!t]  
		\centering
		\includegraphics[width=1\textwidth]{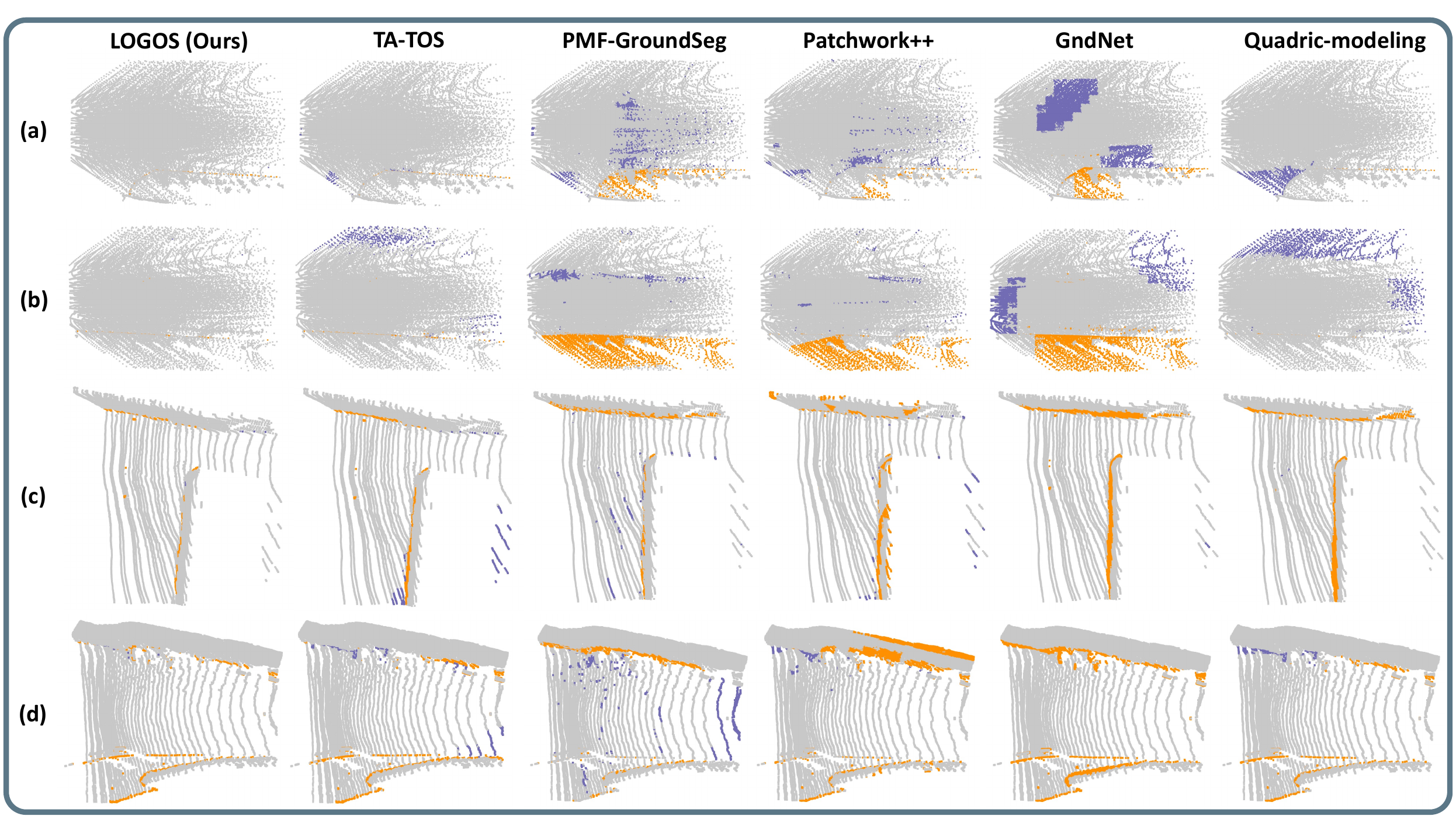}
		\caption{Comparison of our proposed LOGOS with other SOTA tiny obstacle segmentation methods. Rows (a) and (b) show the experimental results in the flat and bumpy scenes on the TOSeg-Road dataset, while rows (c) and (d) shows the experimental results in the flat and bumpy scenes on the TOSeg-Offroad, respectively. \protect\textcolor[RGB]{200, 200, 200}{\rule[0ex]{0.8em}{0.8em}}, \protect\textcolor[RGB]{112, 108, 175}{\rule[0ex]{0.8em}{0.8em}}, and \protect\textcolor[RGB]{247, 150, 32}{\rule[0ex]{0.8em}{0.8em}} refer to the $TP+TN$ result, $FP$ result, and $FN$ result, respectively.}
		\label{fig:exp}
\end{figure*}

To evaluate the proposed LOGOS method, we conduct a series of experiments on the TOSeg-Road dataset and the TOSeg-Offroad dataset \cite{tatos}.

The TOSeg-Road dataset targets urban traffic systems. Data are acquired using a Livox Avia LiDAR (a non-repetitive scanning LiDAR) on structured urban roads. The scenes include both flat and bumpy roads, with a variety of small obstacles such as curbs, manhole covers, sidewalks (positive obstacles), as well as potholes and cracks (negative obstacles).

The TOSeg-Offroad dataset focuses on mining transportation systems. Point clouds are collected using a Falcon-K LiDAR (310-line repeated scanning LiDAR) in an active mining area. The terrain exhibits significant undulations and contains diverse small obstacles such as falling rocks, slopes, wheel ruts, road posts, thin poles, and rock piles.

The two datasets are constructed following an identical pipeline. The source data bags consist of 55 real-world operating sequences captured from distinct scenarios in urban traffic systems and mining transportation systems, each lasting between 30 seconds and 3 minutes. As tiny obstacles occur infrequently in these sequences, a data cleaning process is applied to identify and select only those point cloud frames that contain tiny obstacles, resulting in a total of 1,270 keyframes. These selected keyframes are then manually annotated at the point level, where each point is labeled as “road surface”, “positive obstacle”, or “negative obstacle”. To better evaluate algorithm adaptability to different terrain roughness levels, both TOSeg-Offroad and TOSeg-Road are further divided into two difficulty levels: ``relatively flat'' and ``relatively bumpy''.

\subsubsection{Implementation Details}
\label{sec:implementation}
As discussed in Sec.~\ref{sec:meth:init}, we first employ a sliding-window strategy to initialize the Gaussian primitives. To enhance the adaptability of the sliding-window initialization process to different environments, we set the window size $W$ to twice the stride $S$, ensuring overlap between adjacent windows. Furthermore, since off-road scenes have a larger ROI, a larger window size is adopted. 

Moreover, we set the smoothness threshold $\epsilon_c$ and the elevation threshold $\epsilon_d$. The thresholds $\epsilon_c$ and $\epsilon_d$ are both set to 0.05 m on TOSeg-Road and 0.15 m on TOSeg-Offroad, respectively, as off-road scenes contain more significant terrain undulations. The parameter setting for the proposed LOGOS is shown in Table~\ref{table:param}.

\begin{table}[!ht]
\centering
\setlength\tabcolsep{8pt}
\caption{Parameter setting for our proposed LOGOS.}
\begin{tabular}{cccc}
\toprule[2pt]
 \multirow{2}{*}{Variable} & \multicolumn{2}{c}{Dataset} \\
 &  \;\;\;\;\;TOSeg-Road\;\;\;\;\;        & TOSeg-Offroad    \\ \midrule[1pt]
$S$ &  2 & 6  \\
$W$ &  4 & 12  \\
$\epsilon_c$ &  0.05   &  0.15    \\
$\epsilon_d$ &  0.05   &  0.15    \\
\bottomrule[2pt]
\end{tabular}%
\label{table:param}
\end{table}

\subsubsection{Evaluation Metrics}
\label{sec:metrics}
To quantitatively assess the segmentation performance of different methods, we adopt three widely used metrics: accuracy, F1-score \cite{fan-tip}, and intersection over union (IoU) \cite{iou}. These metrics are defined based on the confusion matrix elements (true positives $TP$, true negatives $TN$, false positives $FP$, and false negatives $FN$) as follows:

\begin{align}
    \text{Accuracy} &= \frac{TP + TN}{TP + TN + FP + FN}, \notag \\
    \text{F1} &= \frac{2 \cdot TP}{2 \cdot TP + FP + FN}, \notag \\
    \text{IoU} &= \frac{TP}{TP + FP + FN}.
    \label{eq:metrics}
\end{align}

In our evaluation on the TOSeg-Road and TOSsg-Offroad datasets, we treat both positive and negative obstacle points as the ``positive'' class, while road surface points are considered the ``negative'' class.

Using these metrics to comprehensively evaluate the effectiveness, robustness, and modular design of the proposed LOGOS, we conduct systematic quantitative and qualitative experiments on the TOSeg-Road and TOSeg-Offroad datasets. The experiments focus on verifying the overall accuracy improvement of LOGOS over the previous SOTA method and emphasizing the improvement on tiny obstacle and in degraded LiDAR scans.

\subsection{Comparative Study on the TOSeg Datasets}
\label{sec:comparative}

\subsubsection{Overall Performance Comparison}
We compare LOGOS with a wide range of representative SOTA approaches on both structured and unstructured scenes. The selected baselines include classification-based method (PMF-GroundSeg \cite{pmf}), fixed parametric modeling methods (Plane-modeling \cite{pm} and Quadric-modeling \cite{qm}), discrete grid modeling methods (Patchwork \cite{patchwork}, Patchwork++ \cite{patchwork++}, DipG-Seg \cite{dipg}, and TA-TOS \cite{tatos}), and regression-based method (GndNet \cite{gndnet}). All methods are evaluated on the TOSeg-Road and TOSeg-Offroad datasets, which are further divided into ``Flat'' and ``Bumpy'' subsets to reflect terrain difficulty. Fig~\ref{fig:exp} shows some examples of the comparative results.

Table~\ref{table:comparison1} shows the results on TOSeg-Road. It can be observed that methods based on simple planar or quadric surface assumptions (e.g., Plane-modeling and Quadric-modeling) perform reasonably well on flat terrain but suffer significant performance drops on bumpy roads. Moreover, plane-based gird methods like Patchwork and Patchwork++ achieve moderate accuracy but undergo errors in some extreme degraded regions, resulting in relatively low F1 and IoU scores. In addition, TA-TOS substantially outperforms all previous methods, achieving an F1 of 0.956 on flat and 0.948 on bumpy subsets. Our LOGOS further improves upon TA-TOS across all metrics: on flat terrain, accuracy rises from 0.989 to 0.991, F1 from 0.956 to 0.966, and IoU from 0.916 to 0.935; on bumpy terrain, F1 increases from 0.948 to 0.957 and IoU from 0.902 to 0.918. These gains demonstrate the advantage of the proposed Gaussian elevation splatting over the MRF-based grid modeling, especially in handling terrain undulations.

Table~\ref{table:comparison2} presents the results on the more challenging TOSeg-Offroad dataset. The performance gap between LOGOS and other methods becomes even more pronounced. While Quadric-modeling and Plane-modeling degrade significantly on bumpy off-road terrain (e.g., Plane-modeling F1 drops to 0.225), TA-TOS maintains high performance (F1 0.939) and LOGOS pushes it further to 0.962. Specifically, on the flat subset of TOSeg-Offroad, LOGOS achieves an F1 of 0.941 (vs. TA-TOS 0.912) and IoU of 0.889 (vs. 0.839). On the bumpy subset, the improvements are even larger: F1 increases from 0.939 to 0.962, and IoU from 0.885 to 0.926. These results confirm that the continuous surface representation and adaptive pruning of LOGOS are particularly beneficial for unstructured, highly undulating environments where discrete grid models often fail.

\begin{table*}[t!]
\centering
\setlength\tabcolsep{10pt}
\caption{Overall performance comparison between tiny obstacle segmentation methods on the \textbf{TOSeg-Road} dataset.  \\The best result is \textbf{bold} and the second-best result is \underline{underlined}.}
\begin{tabular}{llcccccc}
\toprule[2pt]

\multirow{2}{*}{}     & \multirow{2}{*}{Method} & \multicolumn{3}{c}{Flat}              & \multicolumn{3}{c}{Bumpy}                \\
                      &                          & Acc         & F1             & IoU            & Acc     & F1             & IoU            \\ \midrule[1pt]
\multirow{9}{*}{TOSeg-Road} & PMF-GroundSeg \cite{pmf}            &  0.933       &    0.703       &   0.543        &   0.885         &   0.688   &  0.524      \\
& Patchwork \cite{patchwork}            &   0.950  &     0.781      &    0.641       &     0.954       &  0.870  &   0.770       \\
& Patchwork++ \cite{patchwork++}            &   0.951    &   0.808      &    0.678       &    0.932         &   0.822   &    0.697       \\
& DipG-Seg \cite{dipg}            &   0.870   &     0.541        &    0.371       &     0.883        &    0.678    &    0.513       \\
& GndNet [33]           &  0.813   &     0.486        &    0.321       &     0.777       &   0.509  &   0.342   \\
                      & Plane-modeling \cite{pm}         & 0.942     & 0.811 
         & 0.682          & 0.914          &0.798 & 0.664 \\
                      & Quadric-modeling  \cite{qm}       & 0.982    & 0.934          & 0.875          & 0.934          & 0.831          & 0.711    \\
                      & TA-TOS \cite{tatos}   & \underline{0.989}  
                      &  \underline{0.956} & \underline{0.916} & \underline{0.981}   & \underline{0.948} & \underline{0.902} \\ & 
                    \textbf{LOGOS (Ours)}   & \textbf{0.991}  
                      &  \textbf{0.966} & \textbf{0.935} & \textbf{0.984}   & \textbf{0.957} & \textbf{0.918} \\ \bottomrule[2pt]
\end{tabular}%
\label{table:comparison1}
\end{table*}

\begin{table*}[!t]
\centering
\setlength\tabcolsep{10pt}
\caption{Overall performance comparison between tiny obstacle segmentation methods on the \textbf{TOSeg-Offroad} dataset.}
\begin{tabular}{llcccccc}
\toprule[2pt]

\multirow{2}{*}{}     & \multirow{2}{*}{Method} & \multicolumn{3}{c}{Flat}              & \multicolumn{3}{c}{Bumpy}                \\
                      &                          & Acc          & F1             & IoU            & Acc        & F1             & IoU            \\ 
\midrule[1pt]
\multirow{9}{*}{TOSeg-Offroad} 
& PMF-GroundSeg \cite{pmf}    &  0.927    &  0.872       &   0.773     &   0.890    &  0.870   &     0.769   \\
& Patchwork \cite{patchwork}        &   0.910       &    0.835       &    0.717       &    0.860        &    0.803    &  0.671       \\
& Patchwork++ \cite{patchwork++}            &  0.925      &   0.868        &      0.767     &    0.885          &    0.846    &    0.732       \\
& DipG-Seg \cite{dipg}            &   0.502            &     0.524      &    0.355       &     0.547      &    0.602    &   0.430      \\
&  GndNet [33]             &  0.886   &     0.779        &    0.638      &     0.741       &   0.640  &   0.470   \\
                      & Plane-modeling \cite{pm}         & 0.935       & 0.889          & 0.800          & 0.623  & 0.225          & 0.127          \\
                      & Quadric-modeling   \cite{qm}     & 0.930       & 0.886          & 0.795          & 0.740   & 0.635          & 0.465          \\
                      & TA-TOS \cite{tatos}  & \underline{0.948} 
                      & \underline{0.912} & \underline{0.839} & \underline{0.950}  & \underline{0.939} & \underline{0.885} \\ & 
                    \textbf{LOGOS (Ours)}   & \textbf{0.965}
                      &  \textbf{0.941} & \textbf{0.889} & \textbf{0.969}   & \textbf{0.962} & \textbf{0.926} \\ \bottomrule[2pt]
\end{tabular}%
\label{table:comparison2}
\end{table*}

\begin{table}[t!]
\centering
\setlength\tabcolsep{8pt}
\caption{Mid-road tiny \textbf{positive} obstacle segmentation results obtained by methods on the TOSeg datasets.}
\begin{tabular}{llccc}
\toprule[2pt]
&
 \multirow{2}{*}{Method} & \multicolumn{3}{c}{Metric}                  \\  &
  & Acc          & F1             & IoU          \\ \midrule[1pt]
                      \multirow{9}{*}{\rotatebox{90}{Pos. Obs.}} &
    PMF-GroundSeg \cite{pmf}            &  0.905       &    0.310       &   0.184            \\   &
Patchwork \cite{patchwork}            &   0.926        &     0.359      &    0.218          \\    &
Patchwork++ \cite{patchwork++}            &   0.926      &    0.363      &    0.222         \\    &
DipG-Seg \cite{dipg}            &   0.752        &   0.124        &    0.066          \\  &
GndNet \cite{gndnet}            &   0.794        &   0.158        &    0.086    \\ &
                      Plane-modeling \cite{pm}         & 0.862    
         & 0.200        & 0.111          \\   &
                      Quadric-modeling  \cite{qm}       & 0.903         & 0.301          & 0.177           \\  &
                      TA-TOS \cite{tatos}  & \underline{0.972}  
                      &  \underline{0.593} & \underline{0.421}  \\ &
                      \textbf{LOGOS (Ours)}   & \textbf{0.980}  
                      &  \textbf{0.676} & \textbf{0.511} \\
 \bottomrule[2pt]
\end{tabular}%
\label{table:comparison-pos}
\end{table}

\begin{table}[t!]
\centering
\setlength\tabcolsep{8pt}
\caption{Mid-road tiny \textbf{negative} obstacle segmentation results obtained by methods on the TOSeg datasets.}
\begin{tabular}{llccc}
\toprule[2pt]
&
 \multirow{2}{*}{Method} & \multicolumn{3}{c}{Metric}                  \\  &
  & Acc          & F1             & IoU          \\ \midrule[1pt]
                      \multirow{4}{*}{\rotatebox{90}{Neg. Obs.}} &
       Plane-modeling \cite{pm}         & 0.949    
         & 0.048        & 0.024      \\   &
                      Quadric-modeling  \cite{qm}       & 0.977         & 0.094          & 0.049           \\  &
                      TA-TOS \cite{tatos}  & \underline{0.995}  
                      &  \underline{0.302} & \underline{0.178}  \\ &
                      \textbf{LOGOS (Ours)}   & \textbf{0.997}  
                      &  \textbf{0.345} & \textbf{0.209} \\
 \bottomrule[2pt]
\end{tabular}%
\label{table:comparison-neg}
\end{table}

\begin{table*}[t!]
\centering
\setlength\tabcolsep{10pt}
\caption{Performance comparison between LOGOS and TA-TOS on the turning frames.}
\label{table:turning}
\begin{tabular}{lcccccc}
\toprule[2pt]  
                                  & \multicolumn{3}{c}{TA-TOS}                                   & \multicolumn{3}{c}{\textbf{LOGOS (Ours)}} \\
                                  & \multicolumn{1}{c}{Acc} & \multicolumn{1}{c}{F1} & \multicolumn{1}{c}{IoU} & \multicolumn{1}{c}{Acc} & \multicolumn{1}{c}{F1} & \multicolumn{1}{c}{IoU} \\
\midrule[1pt] 
TOSeg-Road         & 0.984                   & 0.951                  & 0.907                   & \textbf{0.987}          & \textbf{0.960}          & \textbf{0.923}          \\
TOSeg-Offroad         & 0.949                   & 0.930                  & 0.870                   & \textbf{0.967}          & \textbf{0.955}          & \textbf{0.914}          \\
TOSeg-Offroad-Turning & 0.880                   & 0.837                  & 0.720                   & \textbf{0.958}          & \textbf{0.940}          & \textbf{0.887}          \\
\bottomrule[2pt]  
\end{tabular}
\end{table*}

\begin{table*}[t!]
\centering
\setlength\tabcolsep{10pt}
\caption{Performance degradation on extremely downsampled \textbf{TOSeg-Road} data.}
\label{table:sparsity-road}
\begin{tabular}{ccccccccc}
\toprule[2pt]
                            & \multicolumn{2}{c}{Sparsity} & \multicolumn{3}{c}{TA-TOS} & \multicolumn{3}{c}{\textbf{LOGOS (Ours)}} \\
                            & $interval$ [m] & $density$ [pts/m$^2$] & Acc   & F1    & IoU   & Acc   & F1    & IoU   \\
\midrule[1pt]
\multirow{6}{*}{TOSeg-Road}   & original & 114.583 & 0.984 & 0.951 & 0.907 & \textbf{0.987} & \textbf{0.960} & \textbf{0.923} \\
                            & 0.3      & 12.058  & 0.976 & 0.948 & 0.901 & \textbf{0.985} & \textbf{0.967} & \textbf{0.936} \\
                            & 0.6      & 3.618   & 0.958 & 0.918 & 0.849 & \textbf{0.974} & \textbf{0.947} & \textbf{0.900} \\
                            & 0.9      & 1.762   & 0.592 & 0.490 & 0.325 & \textbf{0.958} & \textbf{0.912} & \textbf{0.839} \\
                            & 1.2      & 0.901   & 0.084 & 0.146 & 0.079 & \textbf{0.937} & \textbf{0.883} & \textbf{0.791} \\
                            & 1.5      & 0.639   & 0.048 & 0.092 & 0.048 & \textbf{0.946} & \textbf{0.903} & \textbf{0.823} \\
\bottomrule[2pt]
\end{tabular}
\end{table*}

\begin{table*}[t!]
\centering
\setlength\tabcolsep{10pt}
\caption{Performance degradation on extremely downsampled \textbf{TOSeg-Offroad} data.}
\label{table:sparsity-offroad}
\begin{tabular}{ccccccccc}
\toprule[2pt]
                            & \multicolumn{2}{c}{Sparsity} & \multicolumn{3}{c}{TA-TOS} & \multicolumn{3}{c}{\textbf{LOGOS (Ours)}} \\
                            & $interval$ [m] & $density$ [pts/m$^2$] & Acc   & F1    & IoU   & Acc   & F1    & IoU   \\
\midrule[1pt]
\multirow{6}{*}{TOSeg-Offroad} & original & 8.788 & 0.951 & 0.930 & 0.870 & \textbf{0.967} & \textbf{0.955} & \textbf{0.914} \\
                            & 0.5      & 2.066 & 0.933 & 0.889 & 0.801 & \textbf{0.961} & \textbf{0.931} & \textbf{0.871} \\
                            & 1.0      & 0.803 & 0.936 & 0.869 & 0.769 & \textbf{0.964} & \textbf{0.924} & \textbf{0.859} \\
                            & 1.5      & 0.418 & 0.925 & 0.848 & 0.736 & \textbf{0.965} & \textbf{0.925} & \textbf{0.860} \\
                            & 2.0      & 0.253 & 0.307 & 0.318 & 0.189 & \textbf{0.978} & \textbf{0.954} & \textbf{0.913} \\
                            & 2.5      & 0.178 & 0.423 & 0.354 & 0.215 & \textbf{0.962} & \textbf{0.918} & \textbf{0.849} \\
\bottomrule[2pt]
\end{tabular}
\end{table*}

\subsubsection{Performance Comparison for Tiny Obstacle Segmentation}
Tiny obstacles located in the middle of the road pose a significant threat to driving safety, yet they are easily missed by methods that rely on coarse surface fitting. To further demonstrate the superiority of our proposed LOGOS in detecting such small-scale objects, we conduct dedicated experiments focusing on positive obstacles (e.g., rocks, debris) and negative obstacles (e.g., potholes, cracks) that appear in the central driving area. The evaluation follows the same protocol as described in Sec.~\ref{sec:metrics}, but only points within the central ROI are considered.

Table~\ref{table:comparison-pos} summarizes the results for mid-road positive tiny obstacles. Most baseline methods achieve very low F1 scores (below 0.4), indicating that they frequently miss or partially detect small positive obstacles. TA-TOS raises the F1 to 0.593 and IoU to 0.421, already a substantial improvement. Our LOGOS further pushes the F1 to 0.676 and IoU to 0.511, representing a relative gain of 14\% in F1 and 21\% in IoU over TA-TOS. This improvement stems from the BEV-projection elevation splatting module, which preserves fine-grained height variations even for small objects.

Table~\ref{table:comparison-neg} shows the results for negative obstacles (e.g., potholes). This task is extremely challenging because negative obstacles exhibit subtle height drops that can be easily smoothed out by surface models. Plane-modeling and Quadric-modeling achieve near-zero F1 scores (0.048 and 0.094, respectively). TA-TOS achieves an F1 of 0.302 and IoU of 0.178, demonstrating the effectiveness of its MRF-based optimization. LOGOS further improves these to 0.345 and 0.209, respectively. The gain, though modest, is statistically significant given the extreme difficulty of the task. The normal-aware rendering in LOGOS better preserves the signed distance signals of depressions, leading to more accurate detection of negative obstacles.

\subsection{Performance Evaluation on Degraded LiDAR Scans}
One of the primary motivations of LOGOS is to address the performance degradation of SOTA road-modeling-based methods in regions with low point density, such as far-field areas, occluded regions, and rapid turning maneuvers. We evaluate this aspect from two perspectives: (1) performance on challenging turning frames where the LiDAR view changes rapidly, causing sparse and irregular point clouds at scene edges; (2) controlled downsampling experiments that artificially reduce point density to extreme levels.

\subsubsection{Performance on Turning Scenarios}

We select typical frames from the TOSeg-Offroad dataset that involve sharp turns of the mining truck (denoted as TOSeg-Offroad-Turning). These frames exhibit severe edge sparsity and irregular point distribution. Table~\ref{table:turning} compares TA-TOS and LOGOS on these turning frames, along with the full-dataset results for reference.

On the regular TOSeg-Road and TOSeg-Offroad datasets, LOGOS consistently outperforms TA-TOS. More importantly, on the turning subset, TA-TOS suffers a dramatic performance drop: its IoU falls from 0.870 (on full off-road) to 0.720, a degradation of 17\%. In contrast, LOGOS maintains much higher robustness: its IoU only decreases from 0.914 to 0.887, a mere 3\% drop. The absolute IoU of LOGOS on turning frames (0.887) is still higher than the full-dataset IoU of TA-TOS (0.870). These results clearly demonstrate that the sliding-window initialization with adaptive window skipping and the BEV-projection rendering make LOGOS far less sensitive to edge sparsity and irregular point distributions caused by rapid sensor motion.

\subsubsection{Extreme Downsampling Experiments}

\begin{table*}[t!]
\centering
\setlength\tabcolsep{10pt}
\caption{Ablation experiments for LOGOS modules.}
\label{table:ablation-gs}
\begin{tabular}{cccccccc}
\toprule[2pt]
\multicolumn{2}{c}{Module} & \multicolumn{3}{c}{TOSeg-Road} & \multicolumn{3}{c}{TOSeg-Offroad} \\
Pruning & Normal-Aware & Acc & F1 & IoU & Acc & F1 & IoU \\
\midrule[1pt]
          &        & 0.922 & 0.711 & 0.551    & 0.772 & 0.580 & 0.409  \\
$\checkmark$ &     & 0.984 & 0.951 & 0.906       & 0.958 & 0.942 & 0.891  \\
          & $\checkmark$ & 0.920 & 0.692 & 0.529 & 0.657 & 0.176 & 0.096  \\
$\checkmark$ & $\checkmark$  & \textbf{0.987} & \textbf{0.960} & \textbf{0.923} & \textbf{0.967} & \textbf{0.955} & \textbf{0.914}  \\
\bottomrule[2pt]
\end{tabular}
\end{table*}

To systematically quantify robustness to point density, we perform grid downsampling with increasing intervals, generating point clouds with densities ranging from over 100 pts/m$^2$ down to below 0.2 pts/m$^2$. Tables~\ref{table:sparsity-road} and~\ref{table:sparsity-offroad} indicate the performance degradation of TA-TOS and LOGOS on the TOSeg-Road and TOSeg-Offroad datasets, respectively.

On TOSeg-Road (Table~\ref{table:sparsity-road}), TA-TOS maintains reasonable performance down to a density of 1.762 pts/m$^2$ (IoU 0.325), but then collapses catastrophically at 0.901 pts/m$^2$ (IoU 0.079). LOGOS, in contrast, exhibits a graceful degradation: even at the extreme density of 0.639 pts/m$^2$, it still achieves an IoU of 0.823, which is more than ten times higher than IoU of TA-TOS at a comparable density. The continuous Gaussian mixture representation allows LOGOS to infer road elevation from neighboring point-rich regions via the rendering function, whereas grid-wise RANSAC of TA-TOS fails when any grid lacks sufficient points.

On TOSeg-Offroad (Table~\ref{table:sparsity-offroad}), the trend is similar but even more striking. TA-TOS collapses at a density of 0.253 pts/m$^2$ (IoU 0.189), while LOGOS maintains an IoU of 0.913 at the same density — exceeding the original full-density IoU of TA-TOS (0.870). Even at the highest sparsity level (0.178 pts/m$^2$), LOGOS still achieves an IoU of 0.849, demonstrating its exceptional resilience to point cloud sparsity.

These results collectively confirm that LOGOS successfully addresses the core limitation of previous discrete-grid-based methods in low-density regions. The combination of sliding-window initialization (which skips unreliable windows), incremental pruning (which removes obstacle Gaussians), and BEV-projection elevation splatting (which leverages long-range Gaussian influence) provides a robust solution for tiny obstacle segmentation under degraded LiDAR scans.

\subsection{Ablation Study}
\label{sec:ablation}

\begin{table*}[t!]
\centering
\setlength\tabcolsep{10pt}
\caption{Comparison of module computation time between LOGOS and TA-TOS measured in [ms].}
\label{table:time_module}
\begin{tabular}{lcccc}
\toprule[2pt]  
         \multirow{2}{*}{Module}                          & \multicolumn{2}{c}{TOSeg-Road}                                   & \multicolumn{2}{c}{TOSeg-Offroad} \\
                                  & \multicolumn{1}{c}{TA-TOS} & \multicolumn{1}{c}{\textbf{LOGOS (Ours)}}  & \multicolumn{1}{c}{TA-TOS} & \multicolumn{1}{c}{\textbf{LOGOS (Ours)}}  \\
\midrule[1pt] 
Road Model Initialization          & 18                   &    \multirow{2}{*}{\textbf{24}}                           & 7          & \multirow{2}{*}{\textbf{16}}         \\
Road Model Refinement     & 44                  &                             & 27         &          \\
Obstacle Segmentation & \textbf{1}                   & 33                             & \textbf{1}        &       15      \\
\textbf{Total} & 63                 &    \textbf{57}                         & 35         &  \textbf{31}            \\
\bottomrule[2pt]  \\

\end{tabular}
\end{table*}

\begin{table}[t!]
\centering
\setlength\tabcolsep{6pt}
\caption{Computation time comparison for tiny obstacle segmentation measured in [ms].}
\begin{tabular}{lcc}
\toprule[2pt] 
 \multirow{2}{*}{Method} & \multicolumn{2}{c}{Dataset} \\
 &  TOSeg-Road      & TOSeg-Offroad   \\ \midrule[1pt]
 PMF-GroundSeg \cite{pmf} &  5469 & 646 \\
 Patchwork \cite{patchwork} &  \textbf{2} & \textbf{2} \\
 Patchwork++ \cite{patchwork++} &  \underline{4} & \underline{3} \\
 DipG-Seg \cite{dipg} &   5 &  4 \\
GndNet \cite{gndnet} &  26 & 33 \\
Plane-modeling \cite{pm}  &  77   &  32   \\
Quadric-modeling  \cite{qm}  &   84  & 39  \\
TA-TOS \cite{tatos}    &  63   &   35 \\
\textbf{LOGOS (Ours)}    &  57   &   31 \\
\bottomrule[2pt] 
\end{tabular}%
\label{table:time}
\end{table}

To isolate the specific contributions of each core module in LOGOS, we conduct ablation experiments focusing on two key designs: the freespace-aware pruning module and the normal-aware projection (i.e., using normals instead of simple Z-difference). Table~\ref{table:ablation-gs} shows the performance of four configurations on the TOSeg-Offroad and TOSeg-Road datasets.

\subsubsection{Impact of the freespace-aware pruning module} 

Comparing rows 1 and 2 where both Z-coordinate projection is utilized, we observe a dramatic performance improvement brought by the freespace-aware pruning module. On TOSeg-Offroad, pruning raises the F1-score from 0.580 to 0.942 and IoU from 0.409 to 0.891. On TOSeg-Road, F1 increases from 0.711 to 0.951 and IoU from 0.551 to 0.906. Without pruning, all Gaussians generated during initialization are retained for road modeling. These ``obstacle Gaussians'' have means that are typically higher or lower than the true road surface, causing the final rendered reference surface to incorrectly incorporate obstacle regions into the road model. Consequently, many positive and negative obstacle points are misclassified as road, while some true road points may be falsely detected as obstacles because the reference surface is artificially elevated or depressed. Thus, the overall segmentation performance is severely compromised.

When pruning is enabled, the algorithm leverages the terrain smoothness prior to aggressively remove outlier Gaussians that are inconsistent with the already established road model. The retained Gaussian set then more purely and accurately represents the continuously smooth true road surface. As a result, the rendered reference surface is of much higher quality, and segmentation accuracy is greatly improved. This fully demonstrates the critical importance of the proposed incremental pruning strategy for separating road and obstacle signals.

\subsubsection{Comparison of different projection methods} 
Comparing rows 2 and 4 both with pruning, we see the additional gain from normal-aware projection. With pruning ensuring that Gaussians are well-aligned with the road surface, using normal-aware projection yields further improvements over Z-coordinate projection on both datasets (e.g., on TOSeg-Offroad, IoU increases from 0.891 to 0.914). This is because after pruning, the retained Gaussians mostly correspond to true road surfaces, and their computed normals $\mathbf{n}_j$ accurately reflect the local road orientation. On sloped terrain, computing the signed distance as $(\mathbf{p} - \mathbf{g}_j)^\top \mathbf{n}_j$ is more precise than the simple Z-difference $z(\mathbf{p})-z(\mathbf{g}_j)$, thereby enhancing segmentation accuracy.

Comparing rows 1 and 3 both without pruning, we observe an interesting phenomenon: in the absence of pruning, normal-aware projection performs even worse than simple Z-coordinate projection. On TOSeg-Offroad, the F1-score with normal-aware projection drops to as low as 0.176, far below the 0.580 achieved with Z-coordinate projection. The root cause is the disastrous effect of erroneous normals. When pruning is disabled, the Gaussian set is contaminated with many noise Gaussians from obstacles. These obstacle Gaussians yield ``normals'' that are chaotic and can point in arbitrary directions, completely deviating from the true road surface orientation. Under such conditions, applying the normal-aware projection $(\mathbf{p} - \mathbf{g}_j)^\top \mathbf{n}_j$ uses these incorrect $\mathbf{n}_j$ values, leading to physically meaningless projection results.

Therefore, the ablation study clearly demonstrates that the freespace-aware pruning module is the cornerstone of algorithm effectiveness, while the normal-aware projection provides additional refinement on top of a clean road-surface model.

\subsection{Efficiency Evaluation}

To assess the real-time applicability of LOGOS in autonomous driving systems, we measure the average computation time per frame on both TOSeg-Road and TOSeg-Offroad datasets. All experiments are conducted on an Apple Silicon M1 Pro processor. We conduct a module-wise breakdown between TA-TOS and LOGOS in Table~\ref{table:time_module} and a comprehensive comparison with other SOTA methods in Table~\ref{table:time}.

\subsubsection{Module-Wise Comparison}
Table~\ref{table:time_module} decomposes the processing pipeline into three stages: road model initialization, road model refinement (or joint initialization+pruning for LOGOS), and obstacle segmentation. For TA-TOS, initialization (RANSAC-based) takes 18 ms on TOSeg-Road and 7 ms on TOSeg-Offroad, followed by a separate MRF refinement step costing 44 ms and 27 ms, respectively. In LOGOS, the sliding-window initialization and incremental pruning are performed jointly in a single pass, which reduces the overall complexity from a global optimization to a linear incremental process. Consequently, the combined time for incremental Gaussian initialization and pruning in LOGOS is 24 ms on TOSeg-Road and 16 ms on TOSeg-Offroad, which is significantly lower than the sum of the two separate modules in TA-TOS (62 ms and 34 ms, respectively). This demonstrates the efficiency gain of the proposed incremental pruning algorithm.

However, the obstacle segmentation stage of LOGOS is slower than that of TA-TOS: on TOSeg-Road, it increases from 1 ms to 33 ms; on TOSeg-Offroad, from 1 ms to 15 ms. This is because LOGOS uses a continuous Gaussian mixture representation and performs Gaussian elevation splatting (Eq.~\ref{eq:normal_rendering}) for every point, which involves computing weighted contributions from all nearby Gaussians. In contrast, TA-TOS directly reads the precomputed grid height map, which is a simple lookup. Despite the increased cost in the segmentation stage, the overall total runtime of LOGOS is still slightly lower than that of TA-TOS: 57 ms vs. 63 ms on TOSeg-Road, and 31 ms vs. 35 ms on TOSeg-Offroad. The savings in the initialization+refinement stage more than compensate for the additional cost in segmentation.

\subsubsection{Comprehensive Efficiency Comparison with SOTA Methods}
Table~\ref{table:time} compares the total per-frame runtime of LOGOS against other baseline methods. Several observations can be made. First, ground segmentation methods such as Patchwork and Patchwork++ are extremely fast (2–4 ms), as they are designed for efficient general ground extraction. However, as shown in Tables~\ref{table:comparison1} and~\ref{table:comparison2}, these methods lack the capability to accurately segment tiny positive or negative obstacles, especially in complex terrains. In addition, PMF-GroundSeg is an offline method with very high computational cost (over 5 seconds on TOSeg-Road), making it unsuitable for real-time applications. Third, among tiny obstacle segmentation methods, Plane-modeling and Quadric-modeling run in 32–84 ms, which is comparable to TA-TOS (35–63 ms). Our LOGOS achieves total runtimes of 57 ms on TOSeg-Road and 31 ms on TOSeg-Offroad, which are slightly faster than TA-TOS and well within the 100 ms threshold for real-time perception in intelligent transportation systems. The higher runtime on the TOSeg-Road dataset is due to its higher point density, resulting in more processing points per frame.

\section{Conclusion}
\label{sec:con}

This paper presented LOGOS, a LiDAR-only Gaussian-splatting-based tiny obstacle segmentation system that addresses the accuracy limitations of SOTA fixed parametric or discrete grid representations in degraded LiDAR scans by treating the road surface as a continuous mixture of 2D Gaussian primitives. A freespace-aware initialization strategy jointly estimates and prunes road-surface Gaussians without backpropagation, ensuring real-time execution. Moreover, a normal-aware elevation splatting function computes pointwise signed distances using Gaussian normals, eliminating the reliance on RGB data. Experimental evaluations were conducted in the typical road and off-road scenarios. A comparative study against state-of-the-art methods confirmed that LOGOS achieves superior overall accuracy, particularly on tiny obstacles. Dedicated tests on degraded point clouds and challenging turning scenarios demonstrated that our method maintains robust performance where previous approaches undergo a sharp accuracy decline. Ablation studies further validated the individual contributions of the freespace-aware pruning strategy and the normal-aware elevation splatting module. Runtime evaluation further reveals that LOGOS achieves real-time operation capability.

{\small
\bibliographystyle{IEEEtran}
\bibliography{egbib}
\vspace{-50mm}
}

\end{document}